\documentclass[conference]{IEEEtran}
\usepackage{amsmath,amssymb,amsfonts}
\usepackage{graphicx}
\usepackage{textcomp}
\usepackage{xcolor}
\def\BibTeX{{\rm B\kern-.05em{\sc i\kern-.025em b}\kern-.08em
    T\kern-.1667em\lower.7ex\hbox{E}\kern-.125emX}}
\usepackage{balance}  
\usepackage{booktabs} 

\usepackage{float} 
\usepackage{caption}
\usepackage{subcaption}
\usepackage{algorithm,algpseudocode}

\usepackage{amssymb,amsmath,amsthm}
\usepackage{mathtools} 
\usepackage{titlesec}
\allowdisplaybreaks

\newcommand{\new}[1]{{\color{black}#1}}

\begin{document}

\title{Efficient Construction of Nonlinear Models over Normalized Data}

\author{\IEEEauthorblockN{Zhaoyue Cheng}
\IEEEauthorblockA{\textit{University of Toronto}\\
cheng.zhaoyue@cs.toronto.edu}
\and
\IEEEauthorblockN{Nick Koudas}
\IEEEauthorblockA{\textit{University of Toronto}\\
koudas@cs.toronto.edu}

\and
\IEEEauthorblockN{Zhe Zhang}
\IEEEauthorblockA{\textit{York University}\\
zhezhang@yorku.ca}
\and
\IEEEauthorblockN{Xiaohui Yu}
\IEEEauthorblockA{\textit{York University}\\
xhyu@yorku.ca}
}

\maketitle

\begin{abstract}
Machine Learning (ML) applications are proliferating in the enterprise. Relational data which are prevalent in enterprise applications are typically normalized; as a result, data has to be denormalized via primary/foreign-key joins to be provided as input to ML algorithms. In this paper, we study the implementation of popular nonlinear ML models, Gaussian Mixture Models (GMM) and Neural Networks (NN), over normalized data addressing both cases of binary and multi-way joins over normalized relations.

For the case of GMM, we show how it is possible to decompose computation in a systematic way both for binary joins and for multi-way joins to construct mixture models. We demonstrate that by factoring the computation, one can conduct the training of the models much faster compared to other applicable approaches, without any loss in accuracy.

For the case of NN, we propose algorithms to train the network taking normalized data as the input. Similarly, we present algorithms that can conduct the training of the network in a factorized way and offer performance advantages. The redundancy introduced by denormalization can be exploited for certain types of activation functions. However, we demonstrate that attempting to explore this redundancy is helpful up to a certain point; exploring redundancy at higher layers of the network will always result in increased costs and is not recommended.

We present the results of a thorough experimental evaluation, varying several parameters of the input relations involved and demonstrate that our proposals for the training of GMM and NN yield drastic performance improvements typically starting at 100\%, 
which become increasingly higher as parameters of the underlying data vary, without any loss in accuracy.

\end{abstract}
\section{Introduction}
Machine learning (ML) applications in enterprise settings are increasingly becoming mission-critical. As a result, numerous projects both in industry and academia integrate ML techniques in RDBMS, Spark and a variable of other production systems \cite{Polyzotis:2017:DMC:3035918.3054782}. ML algorithms however have been developed on the assumption that data is readily available on a single input source, with the right formats and encodings. Such effective integration of ML technology in critical systems faces a data representation mismatch. Relational data are typically normalized \cite{Beeri:1978:SID:1286643.1286659} and as a result, data has to be denormalized via primary/foreign-key joins and materialized as a single (temporary) relation to be provided as an input to the ML algorithms. Therefore, the learning process commences after joins have been performed on the relations involved.

Consider the example of an analyst modeling customer shopping trends in a store. The analyst builds a model utilizing order details: Orders(OrderID, CustomerID, ItemID, Time, Amount) where ItemID is a foreign key referring to a new table that stores the items sold by the store: Items(ItemID, Price, Size, Colour, Category). A join is required between the two tables because some of the information in the Items table, e.g. Price, Size and Colour are essential features of a predictive model on buying patterns. After the join, the tables are materialized to be a temporary table that has to be provided as an input to the
various ML algorithms utilized for predictive analysis. Examples like this abound in data analytics: in a video streaming company building recommendation models one has to join user viewing history with video information; at a banking application, building fraud detection models or conducting soft customer segmentation requires a join of customer purchasing/spending records with merchant data; in review sites, one has to associate via a join the reviewer meta-data with their reviews for user modeling applications, etc.

Numerous problems arise when building ML models after a primary/foreign-key join. Firstly, normalization removes redundancy in the data, so in the temporary materialized table, redundancy is reintroduced into the table \cite{Beeri:1978:SID:1286643.1286659}. As a result, this imposes increased storage requirements in the temporary materialized table. Moreover, it increases computation costs since redundancy introduces additional computation during the training processes. Essentially computations have to be repeated for each redundant set of attribute values. Furthermore, maintaining the materialized table when the base tables evolve for continuous learning applications introduces unnecessary overheads as well.
Lastly, ML is commonly used for data analytics, which is often an exploratory process utilizing different slices of the data resulting from joining various base tables. Conducting such joins for every exploration is time-consuming and should be avoided \cite{Polyzotis:2017:DMC:3035918.3054782}.

Kumar et al. \cite{DBLP:conf/sigmod/KumarNP15} recognized these issues and proposed specific algorithms to build generalized linear models and execute various linear algebra operations by pushing ML computations through joins to base tables. That way to the extent possible for the specific class of ML models considered, they demonstrated performance advantages in certain scenarios. Yang et al. \cite{svm2020} investigated the case of learning support vector machines with Gaussian kernels over normalized data. We are not aware of any work considering the general case of Gaussian Mixture Models (GMM) \cite{DBLP:books/lib/Murphy12}, various forms of Neural Networks (NN) and Deep Neural Networks (DNN) \cite{DBLP:books/daglib/0040158} executed over normalized input. GMM abound in various modeling tasks; it is an established method to model complex data spaces with diverse multidimensional characteristics. In addition, GMM is prevalent in financial analysis, quantitative finance, astronomy \cite{doi:10.1111/j.1365-2966.2012.21413.x} as well as banking applications especially dealing with the returns of asset classes \cite{DBLP:books/lib/Murphy12}.

For these reasons, we focus on GMM and NN over normalized data sources and investigate the extent to which we can push the computation through the joins to offer performance advantages. To address the problems raised above, firstly, it is essential to reduce the I/O cost by removing the step of materializing the join result due to its large size. Secondly, we have to investigate how to utilize factorization to eliminate redundant calculations to save computation time. From a technical standpoint, the important issue is how to execute these popular ML algorithms while removing redundancy and saving costs without reducing the quality of the outcomes or the scalability of the algorithms.

In this paper, we compare three different applicable algorithms to construct such models. One alternative approach to materializing the join results is to compute the joins in batches on the fly. Our proposed algorithm factorizes the model and pushes the computation through join without redundancy. It not only reduces I/O cost avoiding unnecessary storage but also saves time by eliminating the repeated calculations. Various trade-offs among three algorithms are analyzed. Furthermore,  the proposal is derived for the case of binary join, suitably generalizing to multi-way joins as well. All the algorithms proposed can deliver the correct models with the same quality as the original models thanks to the exact decomposition introduced. 

In particular, for the most general case of GMM, we propose a factorized algorithm named {\em  F-GMM} over data from normalized relations. We use the EM algorithm to train the parameters during E-step and M-step utilizing the reused calculation effectively.
For the case of NN, the proposed algorithm {\em  F-NN} demonstrates the BP algorithm can be decomposed to take normalized data into account. Large benefits can be achieved between the input layer and the first hidden layer during forward and backward propagation. We however do not continue pursuing this at higher layers of the network, because depending on the activation function we can no longer guarantee the exactness of the decomposition.

To our knowledge, this is the first work that deals with the most general form of GMM and NN in the context of normalized data. In summary, our work makes the following contributions: 
\begin{itemize}
 \item For GMM, we present three algorithms ({\em M-GMM}, {\em S-GMM} and {\em F-GMM}) and analyze their performance for binary and multi-way joins. In particular, {\em F-GMM} is a novel technique utilizing the normalized data to explore the reuse of computation.
 \item As for NN, three algorithms ({\em M-NN}, {\em S-NN} and {\em F-NN}) are proposed in a similar way. We explore the opportunity for algorithm {\em F-NN} to use normalized data directly during the forward and backward propagation training phases at the first layer to improve performance. Similarly, we propose solutions for both the binary and multi-way join case.
 \item In addition, for the case of NN, we analyze the impact of different activation functions on sharing computations during the training phase and investigate the computation cost at higher layers of the network.
 \item We present the results of a thorough experimental evaluation testing the impact of dataset parameters on the performance of the proposed algorithms and quantify the benefits. We also utilize publicly available datasets demonstrating impressive performance improvement in real scenarios.
\end{itemize}

This paper is organized as follows. Section \ref{sec:related} reviews related work. In Section \ref{sec:background}, we present background material for GMM and NN as well as the notation required for what follows. Section \ref{sec:problem}, formally defines the problems we focus on in this paper. In Section \ref{sec:gmm}, we present our solution for  GMM followed by section \ref{sec:nn} in which we present our proposed solutions for training NN. In Section \ref{sec:exp}, we present the results of a thorough experimental evaluation of the proposed approaches. Finally, Section \ref{sec:conc} concludes the paper and discusses avenues for future work in this area.

\section{Related Work}
\label{sec:related}

Recently Cheng and Koudas \cite{chen19} presented an algorithm to factorize construction over normalized data focusing on the restricted case of Independent Gaussian Mixture Models(IGMM). The work herein presents a significant generalization to the case of general GMM making no statistical assumptions on the properties of the underlying Gaussians and presents a comprehensive treatment of NN models.

Factorized databases \cite{DBLP:journals/pvldb/BakibayevKOZ13} were proposed with the basic idea to represent relations with join dependencies using algebraically equivalent forms that store less data physically. 

Extending this work, \cite{DBLP:journals/pvldb/KumarJYNP15,sideli,thomas} apply factorized database ideas to ML computation, utilizing primary/foreign-key joins over relations scaling the work to apply in cases where data do not fit in memory. They focus on a general class of linear models (such as logistic regression) which are applicable in certain ML scenarios. Similarly,
\cite{DBLP:conf/sigmod/SchleichOC16} present a factorized framework and demonstrate its applicability to regressions. More specifically, given a feature vector ($x$) in the table generated from  primary/foreign-key joins of two relations ($S$ and $R$), using logistic regression as an example, we need to calculate $w^Tx$ where $w$ is the parameter for the model. Since the table after joins introduces redundancy, factorized learning can reduce the computations in most cases by calculating $ w_S^Tx_S + w_R^Tx_R$ on the relations before executing the join, where $x_S$ and $x_R$ are the features from $S$ and $R$ respectively. This offers performance benefits over approaches that materialize the join results.
Shah et al. \cite{DBLP:journals/pvldb/ShahKZ17} present experimental evidence showing that in some cases avoiding joins has little impact in classification accuracy. Their study however includes datasets in which joins are required for improved accuracy and our work has applicability in all such cases.

In the same general thread, there exist recent works on scalable ML and data mining algorithms \cite{Elgohary:2016:CLA:2994509.2994515, Boehm:2014:HPS:2732286.2732292, mahout_apache_2008,Boehm:2016:SDM:3007263.3007279,Meng:2016:MML:2946645.2946679,Hellerstein:2012:MAL:2367502.2367510,Cai:2014:CPI:2588555.2593680,DBLP:conf/pods/Khamis0NOS18,DBLP:conf/sigmod/Khamis0NOS18}.
The main emphasis of such works is on the efficient implementation of scalable ML algorithms on a data management platform or their effective execution over programming paradigms. We focus on the specific implementation of nonlinear operations used factorization ideas.
Another related research thread includes the implementation of linear algebra systems on data management systems \cite{noauthor_oracle_nodate, 7930004}.
There is increasing interest in building systems with the aim to achieve closer integration of ML with data management \cite{noauthor_oracle_nodate,Boehm:2016:SDM:3007263.3007279}, \cite{Cai:2013:SDM:2463676.2465283,Gao:2017:BLD:3035918.3035937}, \cite{kraska_mlbase:_2013, koc_incrementally_2011,Ewen:2012:SFI:2350229.2350245}. In the context of non-linear models, \cite{rendle_scaling_2013} is concerned with Bayesian Markov Chain Monte Carlo as applied to Factorization Machine (FM) models and \cite{svm2020} present the algorithms for Support Vector Machine (SVM).  In our case, we focus on factorizing the training processes of GMM and NN over normalized data.

\section{Background and Preliminaries}
\label{sec:background}

In this section, we will present the material and notation necessary for the remainder of the paper.

\subsection{Gaussian Mixture Models (GMM) and EM Algorithm}
\label{sec:Gaussian Mixture Models}

GMM \cite{DBLP:books/lib/Murphy12} is a model comprising a fixed number of Gaussian distributions used for data clustering. Assume we are given $N$ training data points $\textbf{x}^{(n)}, 1 \leq n \leq N$ 
of dimension $d$. The distribution of a mixture of $K$ Gaussian components is
$p(\textbf{x}^{n}) = \sum_{k=1}^{K}\pi_kN(\textbf{x}^{(n)}|\mu_k, \Sigma_k)$,
where $\pi_k$ are the mixing coefficients s.t. $\sum_{k=1}^K \pi_k = 1$. The probability density function of the $k^{th}$ Gaussian component of the mixture model is:

{\footnotesize
\begin{align}
N(\textbf{x}^{(n)}|\mu_k, \Sigma_k) = \frac{1}{\sqrt{(2\pi)^d|\Sigma_k|}}e^{-\frac{1}{2}(\textbf{x}^{(n)} - \mu_k)^T \Sigma_k^{-1} (\textbf{x}^{(n)} - \mu_k)}
\label{eq:normal}
\end{align}
}

There are a number of algorithms that can be applied for iteratively training GMM. However, the Expectation-Maximization (EM) algorithm \cite{dempster_maximum_1977} is the most widely used. EM is an iterative method to identify the  maximum likelihood when the model contains an unobserved latent variable. The algorithm iteratively converges to a local minimum. The EM algorithm starts with some initial estimate of the parameters and then iteratively updates the parameters until convergence. Each iteration consists of one E-step and one M-step.

In the E-step, the posterior distribution of the latent variable $z^{(n)}$ for each observation is updated using the current parameters $\mu_k$, $\Sigma_k$ and $\pi_k$.

{\footnotesize
\begin{align}
\gamma_k^{(n)} = p(z^{(n)}=k|\textbf{x}) = \frac{\pi_kN(\textbf{x}^{(n)}|\mu_k,\Sigma_k)}{\sum^K_{j=1}\pi_jN(\textbf{x}^{(n)}|\mu_j, \Sigma_j)}
\label{eq:gamma}
\end{align}}
where $z \sim Categorical(\pi) $. Essentially $\gamma_k^{(n)}$ are our ``soft" guesses for the values of $p(z^{(n)}=k|\textbf{x})$ at this step.

In the M-step, we re-estimate the parameters  using Maximum Likelihood Estimation (MLE) given the current  $\gamma_k^{(n)}$, for all values of $k$ and $n$:

{\footnotesize
\begin{align}
\mu_k =& \frac{1}{N_k}\sum^N_{n=1}\gamma_k^{(n)}\textbf{x}^{(n)}
\label{eq:mu}\\
\Sigma_k =& \frac{1}{N_k}\sum^N_{n=1}\gamma_k^{(n)}(\textbf{x}^{(n)}-\mu_k)(\textbf{x}^{(n)}-\mu_k)^T
\label{eq:sigma}\\
\pi_k =& \frac{N_k}{N} \quad \textrm{with} \quad N_k=\sum^N_{n=1}\gamma_k^{(n)}
\label{eq:pi}
\end{align}
}

GMM is iteratively updated by E-step and M-step until convergence criteria are met.
One of the commonly used criteria for checking convergence is the difference in the following log-likelihood between two iterations is less than a certain threshold.

{\footnotesize
\begin{align}
\ln{ p(\textbf{x}|\pi, \mu, \Sigma) = \sum_{n=1}^N ln(\sum_{k=1}^K \pi_k N(\textbf{x}^{(n)}|\mu_k, \Sigma_k))} \label{eq:a}
\end{align}
}

\subsection{Neural Networks (NN) and BP Algorithm}
\label{sec:Neural Networks}

NN \cite{DBLP:books/lib/Bishop07} is a supervised model which recently gained popularity with the advent of deep learning \cite{DBLP:books/daglib/0040158}. The BP algorithm is the standard method for training NN. It utilizes the current parameters (weights and biases) to compute the error and uses the gradient descent method to propagate the error back to update the parameters. Due to space limitations, we only introduce basic notation and refer the reader to \cite{DBLP:books/lib/Bishop07} for a comprehensive treatment.

A basic NN can be described as a sequence of linear transformations. 
Assuming $ x_i$ where $i \in \{1 \ldots d\}$ is input feature of one instance, we construct linear combinations of the input as:
$a_j = \sum_{i=1}^d w_{ji}^{(1)} x_i + b_j^{(1)}$
where $j \in \{1 \ldots n_h\}$. The superscript (1) denotes the corresponding parameters at the first hidden layer of the network and $n_h$ is the number of the hidden units at this layer. Parameter $w_{ji}^{(1)}$ where $i \in \{1 \ldots d\}, j \in \{1 \ldots n_h\}$ is the weight between the input feature $x_i$ and the hidden unit $h_j$. $b_j^{(1)}$ is the bias for hidden unit $h_j$ in the first hidden layer. 
The value of $a_j$ is transformed via a differentiable activation function $f$ to get $h_j = f(a_j)$ as the output of one hidden unit. 
Examples of $f$  include the Sigmoid function  $\sigma(a) = \frac{1}{1+exp(-a)}$ and the Rectified Linear Unit (ReLU) function $ReLU(a) = max(0,a)$. The outputs of the hidden units at the first layer are combined to synthesize the second hidden layer:
$z_k = \sum_{j=1}^{n_h} w_{kj}^{(2)}h_j + b_k^{(2)}$
where $k \in \{1 \ldots n_l\}$.

\section{Problem description}
\label{sec:problem}

We now introduce formally the problems we focus in this paper. Following the style presented in \cite{DBLP:conf/sigmod/KumarNP15}, there are two relations \textbf{S} ($\underline{SID}$, $X_{S}$, $FK$) and \textbf{R} ($\underline{RID}$, $X_{R}$) with a primary/foreign-key relationship ($\textbf{S}.FK$ refers to $\textbf{R}.RID$), where $X_{S}$ ($X_{R}$) is the name of feature matrix $\textbf{x}_{S}$ ($\textbf{x}_{R}$). We assume that relation \textbf{S} has $n_{S}$ tuples and relation \textbf{R} has $n_{R}$ tuples satisfying $n_{S}>n_{R}$. There are $d_{S}$ features in $n$-th feature vector ${\textbf{x}_{S}^{(n)}}$ and ${d_{R}} = d - d_{S}$ features in ${\textbf{x}_{R}^{(n)}}$. When learning a GMM over the result of the projected equi-join {\footnotesize\textbf{T}($\underline{SID}$, [$X_{S}$ $X_{R}$]) $\leftarrow$ $\pi_{SID,X_{S},X_{R}}$ (\textbf{R} $\bowtie_{RID=FK}$ \textbf{S})}, the feature vector of a tuple in \textbf{T} is the concatenation of the feature vectors from the joining tuples of \textbf{S} and \textbf{R}. For the case of NN, relation \textbf{S} ($\underline{SID}$, $Y$, $X_{S}$, $FK$) has an additional attribute $Y$ which is the target for learning purposes (the projected schema becomes \textbf{T} ($\underline{SID}$, $Y$, [$X_{S}$ $X_{R}$])). Table \ref{tab:1} summarizes the notations used in this paper.

The multi-way join case follows similarly. We are provided $q$ attribute tables $\textbf{R}_i$($\underline{RID_i}, X_{R_i}$), $i \in \{1 \ldots q\}$ and a table \textbf{S}$(\underline{SID}, X_{S}, FK_1, \ldots, FK_q)$ with $q$ foreign keys. We are interested to learn a GMM over the projected equi-join {T} ($\underline{SID}$, [$X_{S}$ $X_{R_1}$ $\ldots$ $X_{R_q}$]) $\leftarrow \pi_{\underline{SID},X_{S},X_{R_1}\ldots X_{R_q}}$ 
({$\textbf R_1$} $\bowtie_{RID_1=FK_1}$ $\ldots$ {$\textbf R_q$} $\bowtie_{RID_q=FK_q}$ \textbf{S}). Similarly, target $Y$ will be added to relation \textbf{S} and \textbf{T} for the training of NN.

Table \textbf{T} introduces redundancy back to the data representation which normalization removed in the first place. Depending on the redundancy introduced trade-offs may exist which we will explore in our ensuing discussion.
 
For purposes of exposition, we assume that the required joins execute in a block nested loops fashion. Our proposals are equally applicable when other types of joins are adopted such as partitioned hash joins.

\begin{table}
\caption{Notations used in the paper \label{tab:1}}
\centering 
\begin{tabular}{|c|c|} \hline
\textbf{Symbol} & \textbf{Meaning}\\ \hline
\textbf{R} & Relation \\ \hline
\textbf{S} & Relation \\ \hline
\textbf{T} & Join result table\\ \hline
$Y$ & Target \\ \hline
$n_{R}$ & Number of tuples in \textbf{R}\\ \hline
$n_{S}$ & Number of tuples in \textbf{S}\\ \hline
$N$ & Number of tuples in \textbf{T} ($N = n_S$)\\ \hline
$d_{R}$ & Number of features in \textbf{R}\\ \hline
$d_{S}$ & Number of features in \textbf{S}\\ \hline
$d$ & Number of features in \textbf{T} ($d = d_R + d_S$)\\ \hline
${\textbf{x}_R}$ &  Feature matrix from \textbf{R}  \\ \hline
${\textbf{x}_S}$ &  Feature matrix from \textbf{S}  \\ \hline
${\textbf{x}}^{(n)}$ & $n$-th feature vector in \textbf{T} \\ \hline
$x_i^{(n)}$ & $i$-th feature in ${\textbf{x}}^{(n)}$ \\ \hline
\end{tabular}
\end{table}

\section{General Gaussian Mixture Model Algorithm}
\label{sec:gmm}

We assume the most general case for GMM with arbitrary covariance matrices. This section starts by introducing the baseline materializing method and an improved algorithm, then proposes the decomposition of the various computations involved that constitutes the basis of our proposal, algorithm {\em  F-GMM}.

{\small
\begin{algorithm}
\caption{Algorithm {\em M-GMM} \label{alg:gmm}}
\begin{algorithmic}[1]
\State Apply join \textbf{S} and \textbf{R} and materialize the table \textbf{T} after join in the database
\Repeat
	\State \textbf{E-step:}
        	\For{$i$ $\leq$ number of batches}
            	\State Read batch $i$ of \textbf{T} into memory
                \State Update responsibility of data points in this batch
                \State $\gamma_k^{(n)} \gets \frac{\pi_kN(\textbf{x}^{(n)}|\mu_k,\Sigma_k)}{\sum^K_{j=1}\pi_jN(\textbf{x}^{(n)}|\mu_j, \Sigma_j)} \qquad \forall n \in$ batch $i$ 
            \EndFor
 	\State \textbf{M-step:}
        	\For{$i$ $\leq$ number of batches}
            \State Read batch $i$ of \textbf{T} into memory
			\State Add to the sum results from this batch
            \State  $Sum_{\mu_k} += \sum^{\text{$|batch ~size ~i|$}}_{n=1}\gamma_k^{(n)}\textbf{x}^{(n)} $
      		\EndFor
            \State Update $\mu_k \gets \frac{1}{N_k} Sum_{\mu_k}$
			\For{$i$ $\leq$ number of batches}
			\State Read batch $i$  of \textbf{T} into memory
			\State Add to the sum results from this batch
            \State $Sum_{\Sigma_k} += \sum^{\text{$|batch ~size ~i|$}}_{n=1}\gamma_k^{(n)}(\textbf{x}^{(n)}-\mu_k)(\textbf{x}^{(n)}-\mu_k)^T $
            \EndFor
            \State Update $\Sigma_k \gets \frac{1}{N_k} Sum_{\Sigma_k}$
		\State Update $\pi_k \gets \frac{N_k}{N}$ with $N_k \gets \sum^N_{n=1}\gamma_k^{(n)}$
\Until Convergence
\end{algorithmic}
\end{algorithm}
}

\subsection{Baseline Approaches}
\label{subsec:ba}

Algorithm {\em  M-GMM} is the baseline solution widely used by analysts currently; it computes the join of the relations involved, materializes the results on the disk and subsequently reads them executing the EM algorithm to complete training. The algorithm is depicted as Algorithm \ref{alg:gmm}.

\begin{figure}[htpb]
\includegraphics[height=7.5cm,width=1\linewidth]{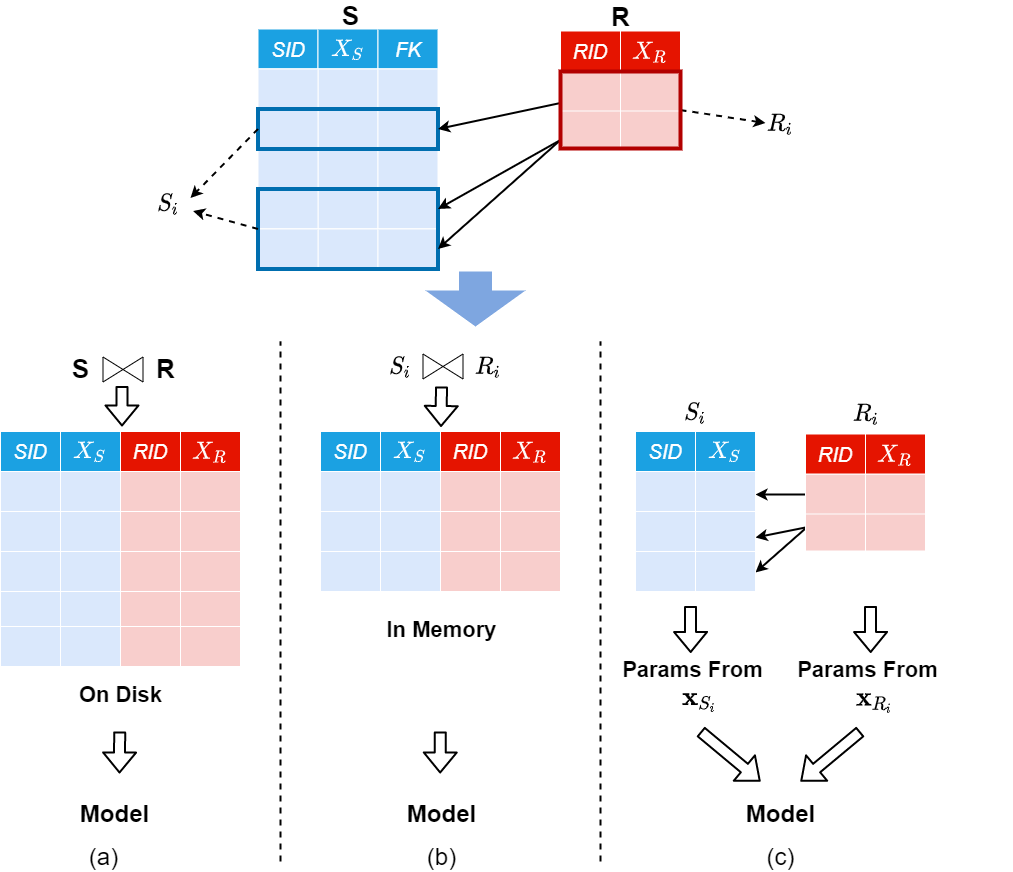}
\caption{\footnotesize \label{fig:example} (a) {\em  M-GMM} (b) {\em  S-GMM} (c) {\em  F-GMM}}
\end{figure}

Algorithm {\em  S-GMM} is another baseline approach that computes the join of the relations involved on the fly without materializing the join result on the disk and executes the EM algorithm directly. This algorithm is essentially the same as Algorithm \ref{alg:gmm}; however, it does not perform Line 1 which is materializing the table \textbf{T} in the database. In addition, in Lines 5, 11, 17, instead of reading $i$-th batch from table \textbf{T},it is accomplished by reading the $i$-th batch of \textbf{R} and retrieving tuples from \textbf{S} using the primary/foreign-key relationship to get the $i$-th sub-table of the join in the memory as the input to compute the GMM on the fly. Figure \ref{fig:example}(a) and (b) present the overview of the two baseline approaches.
{\em  S-GMM} naturally generalizes to multi-way joins.

Here we analyze the cost for the baseline approaches according to Algorithm \ref{alg:gmm}. We use $|S|$, $|R|$, $|T|$ to represent the number of pages in each table. For the case of {\em M-GMM}, the total I/O cost includes computing block nested loops join of  \textbf{S} and \textbf{R} ( $|R|+\frac{|R|}{BlockSize}|S|$), materializing the table \textbf{T} ($|T|$) and executing EM iteratively for $iter$ times where each iteration will read \textbf{T} 3 times ($3\times iter\times |T|$). For the case of {\em S-GMM}, reading \textbf{T} is replaced by executing joins on the fly. Thus, the total I/O cost is
$3\times iter\times (|R|+\frac{|R|}{BlockSize}|S|)$. Which of these two approaches has less I/O cost depends on the size of tables and $BlockSize$. When $BlockSize > \frac{(3\times iter-1)|R||S|}{(3\times iter+1)|T|-(3\times iter-1)|R|}$, {\em S-GMM} has less I/O cost. On the other hand, for each iteration of EM, all the tuples generated from joins must participate in the calculation of each parameter whether they are materialized or not. In other words, there is no difference in terms of computation cost between them. 

\subsection{Algorithm F-GMM for Binary Joins}

\label{subsec:fgmm}
Algorithm {\em  F-GMM} computes the parameters of GMM in a factorized way.
The algorithm is derived from  {\em  M-GMM} and {\em  S-GMM} as follows. As in {\em S-GMM}, materializing the table (Line 1) is not  required; for Lines 5, 11, 17, relation \textbf{R} is processed in batches, probing \textbf{S} for matching tuples using the primary/foreign key. The difference from {\em S-GMM} is that, instead of feeding every joined tuple to the network for parameter updates in Lines 7, 13, 19, we factorize the equations into two parts involving ${\textbf{x}_{S}}$ and ${\textbf{x}_{R}}$ respectively. As illustrated by Figure \ref{fig:example}(c), the $i$-th batch of \textbf{R} ($R_i$) is only used in probing \textbf{S} to identify $S_i$; the actual computation and storage of the join result does not take place. The updated values are then independently computed for those parameters involving $\textbf x_{S_i}$ and those involving $\textbf x_{R_i}$. More specifically, we factorize the computation of $\gamma_k^{(n)}$, $\mu_k$ and $\Sigma_k$ in the E-step as well as the M-step. Some bookkeeping during the factorization is required and provided below.

\subsubsection{E-step}
\label{subsec:f-gmm-e}
In the E-step, $\gamma_k^{(n)}$ needs to be calculated involving Equations (\ref{eq:normal}) and (\ref{eq:gamma}).
In Equation (\ref{eq:normal}), feature vectors are not directly involved in the calculation of $\frac{1}{\sqrt{(2\pi)^d|\Sigma_k|}}$.
In contrast, for the part of $e^{-\frac{1}{2}(\textbf{x}^{(n)} - \mu_k)^T\Sigma_k^{-1}(\textbf{x}^{(n)} - \mu_k)}$, feature vectors from both table \textbf{S} and \textbf{R} are required when computing $(\textbf{x}^{(n)}-\mu_k)^T\Sigma^{-1}_k(\textbf{x}^{(n)}-\mu_k)$. 

To ease notation, we denote $\Sigma_k^{-1}$ as $I_k$ and ignore the subscript k in the expanded equations for all the parameters belonging to the $k^{th}$ Gaussian component.

{\footnotesize
\begin{align}
\underset{1 \times 1}{(\textbf{x}^{(n)}-\mu_k)^TI_k(\textbf{x}^{(n)}-\mu_k)} &= \notag\\
\underset{1 \times d}{\begin{bmatrix} x_1^{(n)}-\mu_1 & x_2^{(n)}-\mu_2 & \cdots & x_d^{(n)}-\mu_d \end{bmatrix}}
&\times \notag \\
\underset{d \times d}
{\begin{bmatrix} 
  I_{1,1} & I_{1,2} & \cdots & I_{1,d}\\
  I_{2,1} & I_{2,2} & \cdots & I_{2,d} \\
  \vdots  & \vdots  & \ddots & \vdots  \\
  I_{d,1} & I_{d,2} & \cdots & I_{d,d}
\end{bmatrix}}
&\times
\underset{d \times 1}
{\begin{bmatrix}
x_1^{(n)}-\mu_1 \\
x_2^{(n)}-\mu_2 \\
\vdots \\
x_d^{(n)}-\mu_d
\end{bmatrix}}
\label{eq:estep}
\end{align}
}
Denote the first $d_S$ dimensions of vector $\textbf{x}^{(n)} - \mu_k $ as $PD_{S}$ and the remaining $d_R$ dimensions  as $PD_R$:

{\footnotesize
\begin{align}
\underset{d_S \times 1}{PD_S} = 
\begin{bmatrix}
x_1^{(n)}-\mu_1 \\
x_2^{(n)}-\mu_2 \\
\vdots \\
x_{d_S}^{(n)}-\mu_{d_S}
\end{bmatrix}& \quad 
\underset{d_R \times 1}{PD_R} = 
\begin{bmatrix}
x_{d_S+1}^{(n)}-\mu_{d_S+1} \\
x_{d_S+2}^{(n)}-\mu_{d_S+2} \\
\vdots \\
x_{d}^{(n)}-\mu_{d}
\end{bmatrix}
\end{align}
}
Then $(\textbf{x}^{(n)}-\mu_k)^TI_k(\textbf{x}^{(n)}-\mu_k)$ = \textbf{UL} (upper left matrix) + \textbf{UR} (upper right matrix) + \textbf{LL} (lower left matrix) + \textbf{LR} (lower right matrix), where

{\footnotesize
\begin{align}
\underset{1 \times 1}{\textbf{UL}} &= \underset{1 \times d_S}{PD_S^T} \quad
\underset{d_S \times d_S}{
  \begin{bmatrix}
  I_{1,1}  & \cdots & I_{1,d_S}\\
  \vdots   & \ddots & \vdots \\
  I_{d_S,1}  & \cdots & I_{d_S,d_S}
 \end{bmatrix}}
 \quad
\underset{d_S \times 1}{PD_S} 
\label{eq:ul}\\
\underset{1 \times 1}{\textbf{UR}} &= \underset{1 \times d_S}{PD_S^T} \quad
\underset{d_S \times d_R}{
  \begin{bmatrix}
  I_{1,d_S+1}  & \cdots & I_{1,d}\\
  \vdots   & \ddots & \vdots \\
  I_{d_S,d_S+1}  & \cdots & I_{d_S,d}
 \end{bmatrix}} 
 \quad
\underset{d_R \times 1}{PD_R}
\label{eq:ur}\\
\underset{1 \times 1}{\textbf{LL}}  &= \underset{1 \times d_R}{PD_R^T} \quad
\underset{d_R \times d_S}{
  \begin{bmatrix}
  I_{d_S+1,1}  & \cdots & I_{d_S+1,d_S}\\
  \vdots   & \ddots & \vdots \\
  I_{d,1}  & \cdots & I_{d,d_S}
 \end{bmatrix}} \quad
\underset{d_S \times 1}{PD_S} 
\label{eq:ll}\\
\underset{1 \times 1}{\textbf{LR}}  &= \underset{1 \times d_R}{PD_R^T} \quad
\underset{d_R \times d_R}{
  \begin{bmatrix}
  I_{d_S+1,d_S+1}  & \cdots & I_{d_S+1,d}\\
  \vdots   & \ddots & \vdots \\
  I_{d,d_S+1}  & \cdots & I_{d,d}
 \end{bmatrix}} \quad
\underset{d_R \times 1}{PD_R}
\label{eq:lr}
\end{align}
}

We are decomposing a multiplication of three matrices in Equation (\ref{eq:estep}) into a sum of four values (Equations (\ref{eq:ul}), (\ref{eq:ur}), (\ref{eq:ll}), (\ref{eq:lr})) each of which is the result of a multiplication of three smaller matrices. They are the intermediate products that do not require every entire tuple $\textbf{x}^{(n)}$ to be calculated. Instead, all of them are calculated using  $\textbf{x}_S^{(n)}$ and $\textbf{x}_R^{(n)}$ from \textbf{S} and \textbf{R} directly. In particular, ${PD_R}$ in Equation (\ref{eq:ur}) and (\ref{eq:ll}) as well as {\textbf{LR}} in Equation (\ref{eq:lr}) only involve $\textbf{x}_R^{(n)}$. Their values can be reused because each tuple in \textbf{R} can match several tuples in \textbf{S} through a primary/foreign-key probing. Therefore, there is a potential for large savings in this way of decomposition by removing the repeated calculations.

\subsubsection{M-step}
\label{subsec:f-gmm-m}
In the M-step, we can update $\pi_k$ using Equation (\ref{eq:pi}) directly
which does not involve the data from table \textbf{S} or \textbf{R}.
When updating $\mu_k$, Equation (\ref{eq:mu}) can be decomposed into two  parts:

{\footnotesize
\begin{align}
\underset{d \times 1}{\mu_k} &= \frac{1}{N_k}\sum^N_{n=1}\gamma_k^{(n)}\textbf{x}^{(n)}
= \begin{bmatrix} 
\underset{d_S \times 1}{\frac{1}{N_k}\sum^N_{n=1}\gamma_k^{(n)}{\textbf{x}_S^{(n)}}} \\
\underset{d_R\times 1}{\frac{1}{N_k}\sum^N_{n=1}\gamma_k^{(n)}{\textbf{x}_R^{(n)}}} 
\end{bmatrix}
\label{eq:mmu}
\end{align}
}

We next outline how to update $\Sigma_k$ using Equation (\ref{eq:sigma}).
We only need to focus on the computation of ${(\textbf{x}^{(n)}-\mu_k)(\textbf{x}^{(n)}-\mu_k)^T}$ since multiplying by a constant $\gamma_k^{(n)}$ and summing over all data points can be easily accomplished as long as we determine how to compute it in a factorized way. Subscript $k$ is also ignored for simplifying notation. 

{\footnotesize
\begin{align}
\underset{d \times d}{(\textbf{x}^{(n)}-\mu_k)(\textbf{x}^{(n)}-\mu_k)^T} &= \notag\\
\underset{d \times 1}
{\begin{bmatrix}
x_1^{(n)}-\mu_1 \\
x_2^{(n)}-\mu_2 \\
\vdots \\
x_d^{(n)}-\mu_d
\end{bmatrix}} \times&
\underset{1 \times d}{\begin{bmatrix} x_1^{(n)}-\mu_1 & x_2^{(n)}-\mu_2 & \cdots & x_d^{(n)}-\mu_d \end{bmatrix}} 
 \notag\\
=& \begin{bmatrix}
\underset{d_S \times d_S}{\textbf{UL}} & \underset{d_S \times d_R}{\textbf{UR}} \\
\underset{d_R \times d_S}{\textbf{LL}} & \underset{d_R \times d_R}{\textbf{LR}}
\label{eq:msigma}
\end{bmatrix}
\end{align}
}
where 
{\footnotesize
\begin{align}
\label{eq:mul}
\underset{d_S \times d_S}{\textbf{UL}} &= 
\underset{d_S \times 1}{PD_S} \quad
\underset{1 \times d_S}{PD_S^T} \\
\label{eq:mur}
\underset{d_S \times d_R}{\textbf{UR}} &= 
\underset{d_S \times 1}{PD_S} \quad
\underset{1 \times d_R}{PD_R^T} \\
\label{eq:mll}
\underset{d_R\times d_S}{\textbf{LL}} &= 
\underset{d_R \times 1}{PD_R} \quad
\underset{1 \times d_S}{PD_S^T} \\
\label{eq:mlr}
\underset{d_R\times d_R}{\textbf{LR}} &=
\underset{d_R \times 1}{PD_R} \quad
\underset{1 \times d_R}{PD_R^T}
\end{align} 
}

In this case, similar to the calculation of $\gamma_k^{(n)}$, all the sub-matrices in Equations (\ref{eq:mmu}) and (\ref{eq:msigma}) can be updated directly from the feature vectors of \textbf{S} and \textbf{R}. Equations (\ref{eq:mul}), (\ref{eq:mur}), (\ref{eq:mll}), (\ref{eq:mlr}) are four matrices with smaller dimensions where ${PD_R}$ and $\textbf{LR}$ can be computed only once and then reused for any tuple in the join result with same $\textbf{x}_R^{(n)}$. The potential for great savings which will be verified in Section \ref{sec:exp} depends on the degree of redundancy introduced.

We take Equation (\ref{eq:msigma}) as an example to analyze the computational savings due to factorization and how redundancy influences its performance. Before decomposition, computing one tuple in \textbf{T} requires $d$ subtractions and $d^2$ multiplications. For $N$ tuples, the original computation time is represented as  $\tau = Nd (\tau_s + d \tau_m)$ where $\tau_s$ and $\tau_m$ represent the time required for one subtraction and multiplication operation respectively. When ${PD_R}$ and $\textbf{LR}$ are reused after decomposition by  {\em F-GMM}, it requires $n_Sd_S+n_Rd_R$ subtractions and $n_S(d_S^2 + 2d_Sd_R) + n_Rd_R^2$ multiplications. As $N = n_S$ and  $d = d_S + d_R$, the time saving is $\Delta \tau = (n_S - n_R) d_R (\tau_s + d_R \tau_m)$ and the saving rate is $\frac{\Delta \tau}{\tau}=\frac{(\frac{n_S}{n_R}-1)(\tau_s+d_R \tau_m)}{\frac{n_S}{n_R}(\frac{d_S}{d_R}+1)(\tau_s+d \tau_m)}$. That is, when $d_S$ is fixed, with the increase of $d_R$ or $\frac{n_S}{n_R}$, the factorized algorithm enjoys more computational cost savings over the baseline algorithms. Moreover, {\em F-GMM} has the same I/O cost as {\em S-GMM}, which we have analyzed in  \ref{subsec:ba}.

Notice that the correctness of the calculation can be guaranteed for the reason that it is an exact decomposition to covert the large matrix operation into several small parts and no approximation is involved. 
All the other factors, such as the input data, the calculation of parameters, and the number of iterations required for training, remain unchanged.
Although {\em  M-GMM} reads the $i$-th batch from \textbf{T} and {\em  S-GMM} / {\em  F-GMM} reads the $i$-th batch from \textbf{R} for probing and then training, the values of parameters updated in each iteration are the same. This is because all $N$ tuples are involved in calculating the parameters in Lines 7, 13, and 19, regardless of the number of matching tuples in each batch. Thus, the parameters are the same after finishing training and the accuracy of the models will not change for the algorithms {\em M-GMM}, {\em S-GMM} and {\em F-GMM}.

\subsection{Algorithm F-GMM for Multi-way Joins}
\label{sec:multigmm}

For large warehouses, multi-way joins are the norm and such a generalization is imperative. We now present the generalization of {\em  F-GMM} to multi-way joins. In this section, we perform the factorization over a join sequence
$\textbf{R}_1$ $\bowtie_{RID_1 = FK_1}$ $\ldots$ $\textbf{R}_q$ $\bowtie_{RID_q = FK_q}$ $\textbf{S}$. To ease notation in what follows, we denote $\textbf{S}$ as $\textbf{R}_0$ and $d_S$ as $d_{R_0}$.

\subsubsection{E-step}

Similarly to the binary case, the E-step involves Equations (\ref{eq:normal}) and (\ref{eq:gamma}).
In contrast, in the calculation of $e^{-\frac{1}{2}(\textbf{x}^{(n)} - \mu_k)^T\Sigma_k^{-1}(\textbf{x}^{(n)} - \mu_k)}$, data from tables \textbf{S} ($\textbf{R}_0$)  to $\textbf{R}_q$ are required when computing $(\textbf{x}^{(n)}-\mu_k)^T\Sigma^{-1}_k(\textbf{x}^{(n)}-\mu_k)$ in Equation (\ref{eq:estep}). 

Denote the first $d_{R_0}$ dimensions of vector $\textbf{x}^{(n)}$ - \textbf{$\mu_k$} as \textbf{$PD_{R_0}$} and the remaining dimensions can be decomposed into $q$ parts corresponding to relations $\textbf{R}_1$ to $\textbf{R}_q$. They are denoted as \textbf{$PD_{R_i}$} with dimension $d_{R_i}$ for $i = 0 \ldots q$. Then

{\footnotesize 
\begin{align}
&\underset{1 \times 1}{(\textbf{x}^{(n)}-\mu_k)^T I_k(\textbf{x}^{(n)}-\mu_k)} = \sum_{i=0}^q \sum_{j=0}^q {PD_{R_i}^T  I_{ij}  PD_{R_j}} 
\label{eq:M-gamma}
 \end{align}

where \begin{align}
\underset{d_{R_m} \times 1}{PD_{R_m}} =& 
{\begin{bmatrix}
x_{d_{m-1}+1}^{(n)}-\mu_{d_{m-1}+1} \\
x_{d_{m-1}+2}^{(n)}-\mu_{d_{m-1}+2}\\
\vdots \\
x_{d_{m}}^{(n)}-\mu_{d_{m}}
\end{bmatrix}}
\label{eq:M-pdrm1}
\\
 \underset{d_{R_m} \times d_{R_n}}{I_{mn}} =&
  \begin{bmatrix}
  I_{d_{m-1}+1,d_{n-1}+1}  & \cdots & I_{d_{m-1}+1,d_{n}}\\
  \vdots   & \ddots & \vdots \\
  I_{d_{m},d_{n-1}+1}  & \cdots & I_{d_{m},d_{n}}
 \end{bmatrix} 
 \end{align}
}

$I_{mn}$ is the partial $I_k$ corresponding to relations $R_m$ and $R_n$, where $m \in \{1 \ldots q\}, n\in \{1 \ldots q\}$ and $d_m = \sum_{i=0}^m {d_{R_i}}$.

In Equation (\ref{eq:M-gamma}), $(\textbf{x}^{(n)}-\mu_k)^T I_k(\textbf{x}^{(n)}-\mu_k)$ is decomposed into a sum of  $(q+1) \times (q+1)$ smaller matrices. Similar to the analysis for the binary join case, when $i=j$ and $i\not = 0$, $PD_{R_i}^T  I_{ij}  PD_{R_j}$ can also be reused due to the redundancy. Furthermore, $\forall m\in\{1 \ldots q\}$, for each feature vector in $\textbf{R}_m$, Equation (\ref{eq:M-pdrm1}) is only calculated once to removing the repeated calculation in {\em S-GMM}.

\subsubsection{M-step}
In the M-step, when combining the features in table \textbf{S} ($\textbf{R}_0$) and tables $\textbf{R}_1$ to $\textbf{R}_q$, we can  decompose $\mu_k$ in Equation (\ref{eq:mu}) as follows:

{\footnotesize
\begin{align}
\underset{d \times 1}{\mu_k} &=\frac{1}{N_k}\sum^N_{n=1}\gamma_k^{(n)}\textbf{x}^{(n)} 
= \begin{bmatrix} 
\underset{d_{R_0} \times 1}{\frac{1}{N_k}\sum^N_{n=1}\gamma_k^{(n)}{\textbf{x}_{R_0}^{(n)}}}\\ 
\underset{d_{R_1} \times 1 }{\frac{1}{N_k}\sum^N_{n=1}\gamma_k^{(n)}{\textbf{x}_{R_1}^{(n)}}}\\
\vdots\\
\underset{d_{R_q} \times 1 }{\frac{1}{N_k}\sum^N_{n=1}\gamma_k^{(n)}{\textbf{x}_{R_q}^{(n)}}}
\end{bmatrix}
\label{eq:M-mu}
\end{align}
}

Next, for updating $\Sigma_k$, Equation(\ref{eq:msigma}) for multi-way joins can be written as:

{\footnotesize
\begin{align}
\begin{bmatrix}
 \underset{d_{R_0} \times d_{R_0}}{M_{00}} & \underset{d_{R_0} \times d_{R_1}}{M_{01}} & \cdots & \underset{d_{R_0} \times d_{R_q}}{M_{0q}} \\
\vdots & \vdots & \ddots & \vdots \\
\underset{d_{R_q} \times d_{R_0}}{M_{q0}} & \underset{d_{R_q} \times d_{R_1}}{M_{q1}} & \cdots & \underset{d_{R_q} \times d_{R_q}}{M_{qq}} 
\end{bmatrix}
\label{eq:M-sigma}
\end{align} 
}
where
{\footnotesize
\begin{align}
\label{eq:42}
\underset{d_{R_i} \times d_{R_j}}{M_{ij}} &= 
\underset{d_{R_i} \times 1}{PD_{R_i}} \quad
\underset{1 \times d_{R_j}}{PD_{R_j}^T} 
\end{align}
} 

In this step, we decompose the result of $(\textbf{x}^{(n)}-\mu_k)(\textbf{x}^{(n)}-\mu_k)^T$ with dimension $d \times d$ into $(q+1)\times(q+1)$ blocks of much smaller matrices. It is evident that there are large savings if we reuse the computation of $PD_{R_i}$ ($i \not= 0$) and $M_{ij}$ ($i=j$ and $i\not = 0$) by getting the features from tables \textbf{S} and $\textbf{R}_1$ to $\textbf{R}_q$ directly for the factorized Equation (\ref{eq:M-sigma}).

\section{Neural Networks (NN)}
\label{sec:nn}

The baseline approach to train a NN, with input from table \textbf{T} on the disk, is referred to as {\em M-NN}. When joining \textbf{R} and \textbf{S} on the fly and feeding the sub-tables to the model without materializing the result, it yields the other baseline algorithm {\em S-NN}. {\em F-NN} is the factorized approach we proposed.
Training the NN can proceed according to standard algorithms namely batch, mini-batch or stochastic gradient descent (SGD) \cite{DBLP:books/lib/Bishop07}. Note that for the case of SGD training, the join of \textbf{S} and \textbf{R} has to be permuted per epoch. Similarly, to perform SGD training in algorithm {\em S-NN} and {\em F-NN}, we can permute the keys of \textbf{R} for each training epoch, accessing the keys in a different order per epoch while probing relation \textbf{S}. Thus, the entire discussion that follows applies equally to mini-batch, batch and SGD training. The detailed decomposition for the computation process in the training process of NN using {\em F-NN} is discussed below.

\subsection{ Algorithm F-NN for Binary Joins}
\label{subsec:fnn}

 \subsubsection{Forward Propagation in the First Layer}\label{sec:ff}

\begin{figure}[htb]
\includegraphics[height=5.0cm, width=1\linewidth]{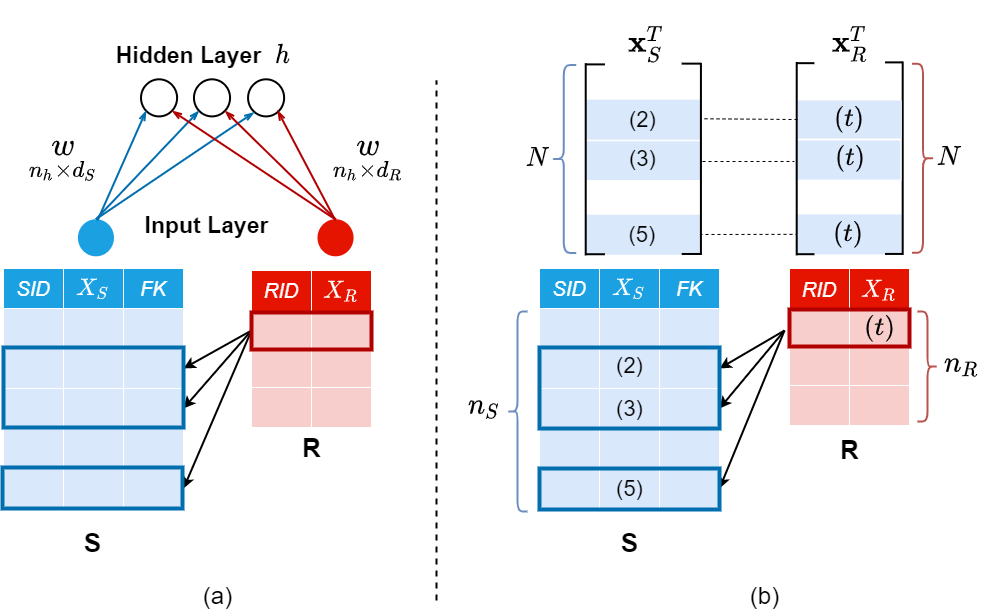}
\caption{ \label{fig:examplenn}
(a) Forward Propagation
(b) Backward Propagation}
\end{figure}

Assume there is one NN that receives $d$ features in the input layer and  have $n_h$ hidden units in the first hidden layer $h$. When receiving the $n$-th feature vector $\textbf{x}^{(n)}$, the value for a single hidden unit in layer $h$ before applying the activation function $f$ is: $a_j^{(n)}  = \sum_{i=1}^d w_{ji}^{(1)} x_i^{(n)} + b_j^{(1)}$ where $j \in \{1 \ldots n_h\}$. In order to make it clearer, we use the form of matrix to present the decomposition process. Thus, the weights matrix between the input and hidden layer is ${w}$ and the bias vector is $b$. To ease notation, we eliminate the superscript (1) denoting the layer index. The vector $a^{(n)}$ for all $a_j^{(n)}$ can be written as:

{\footnotesize
 \begin{align}
\underset{n_h \times 1 } {a^{(n)}} &=  \underset{n_h \times d }{w} \times \underset{d \times 1} {\textbf{x}^{(n)}} + \underset{n_h \times 1}{b}\notag \\
&= \underset{n_h \times d}{
\begin{bmatrix} 
w_{1, 1} & w_{1, 2} & \cdots & w_{1, d} \\
w_{2, 1} & w_{2, 2} & \cdots & w_{2, d} \\
\vdots & \vdots & \ddots & \vdots \\
w_{n_h, 1} & w_{n_h, 2} & \cdots & w_{n_h, d} \\
\end{bmatrix}} 
 \times \underset{d \times 1}{\begin{bmatrix} x_1^{(n)}\\ x_2^{(n)}\\ \vdots\\  x_d^{(n)}\end{bmatrix}} 
+ \underset{n_h \times 1}{\begin{bmatrix} b_1 \\ b_2\\ \vdots\\b_{n_h} \end{bmatrix}} \notag\\
&=\underset{n_h \times d_S}{
\begin{bmatrix} 
w_{1, 1} & w_{1, 2} & \cdots & w_{1, d_S} \\
w_{2, 1} & w_{2, 2} & \cdots & w_{2, d_S} \\
\vdots & \vdots & \ddots & \vdots \\
w_{n_h, 1} & w_{n_h, 2} & \cdots & w_{n_h, d_S} \\
\end{bmatrix}} 
 \times \underset{d_S \times 1}{\begin{bmatrix} x_1^{(n)} \\ x_2^{(n)} \\\vdots\\  x_{d_S}^{(n)}\end{bmatrix}} \notag\\
&+ 
 \underset{n_h \times d_R}{
\begin{bmatrix} 
w_{1, d_{S}+1} & w_{1, d_{S}+2} & \cdots & w_{ 1, d} \\
w_{2, d_{S}+1} & w_{2, d_{S}+2,} & \cdots & w_{2, d} \\
\vdots & \vdots & \ddots & \vdots \\
w_{n_h,d_{S}+1} & w_{n_h, d_{S}+2} & \cdots & w_{n_h, d} 
\end{bmatrix}}
 \times \underset{d_R\times 1}{\begin{bmatrix} x_{d_S+1}^{(n)} \\ x_{d_S+2}^{(n)}\\\vdots \\ x_{d}^{(n)}\end{bmatrix}}\notag
+ \underset{n_h \times 1}{\begin{bmatrix} b_1 \\b_2\\ \vdots \\ b_{n_h} \end{bmatrix}} 
\end{align}
}
In the computation above, the features can be divided into two vectors from relation \textbf{S} and \textbf{R} respectively. After adding the bias vector to apply activation function on the result, we obtain the values of the hidden units $h^{(n)} = f(a^{(n)})$ in the first layer. As depicted in Figure \ref{fig:examplenn} (a),  we can read tuples directly by batches from \textbf{R} and probe \textbf{S} for matching tuples so that parts of the computation of the vector can be pushed to the activation function. For each iteration, the values of the weights and biases are constant and each tuple in \textbf{R} may match several tuples in \textbf{S}. The result of partial inner products involved the features from \textbf{R} adding the bias only needs to be calculated once. In other words, $\sum_{i=d_S+1}^{d} w_{ji}^{(1)} x_i^{(n)} + b_j^{(1)}$ where $j \in \{1 \ldots n_h\}$ is the reused calculation which can bring calculation savings in {\em F-NN} and we will quantify the benefits in Section \ref{sec:exp}.
 
 \subsubsection{Forward Propagation in the Second Layer and Beyond}
Let us now observe the computation between the first layer $h$ and the second layer $l$ with $n_l$ hidden units. For a single unit $l_k$ in the second layer, where $k \in \{1 \ldots n_l\}$, the output after the activation function $f$ is:

{\footnotesize
\begin{align}
l_k^{(n)} &= f(\sum_{j=1}^{n_h} w_{kj}^{(2)}h_j^{(n)} + b_k^{(2)}) 
\label{eq:b0}\\
 &= f(\sum_{j=1}^{n_h} w_{kj}^{(2)}f(\sum_{i=1}^d w_{ji}^{(1)}x_i^{(n)} + b_j^{(1)}) + b_k^{(2)}) 
\end{align}
}

If $f$ is an additive function\footnote{A solution to the Cauchy functional form $f(x+y) = f(x) + f(y)$.}, $l_k^{(n)} $ can be factored into:
{\footnotesize
\begin{align} 
 &f(\sum_{j=1}^{n_h}( w_{kj}^{(2)}f(\sum_{i=1}^{d_S} w_{ji}^{(1)}x_i^{(n)}) + w_{kj}^{(2)}f(\sum_{i=d_S+1}^d w_{ji}^{(1)}x_i^{(n)} + b_j^{(1)}))+ b_k^{(2)})\notag\\
  \label{eq:b}
 &= f(\sum_{j=1}^{n_h} w_{kj}^{(2)}f(T_1) + \sum_{j=1}^{n_h} w_{kj}^{(2)}f(T_2) + b_k^{(2)})
\end{align}
}
where $T_1 = \sum_{i=1}^{d_S} w_{ji}^{(1)}x_i^{(n)}$ and $T_2 = \sum_{i=(d_S+1)}^d w_{ji}^{(1)}x_i^{(n)} + b_j^{(1)}$, which can be stored after being computed in the first layer and reused for the second layer to save additional operations. Moreover, let $T_3 = \sum_{j=1}^{n_h} w_{kj}^{(2)}f(T_2)+b_k^{(2)}$, which is the sum of partial output involved the features from \textbf{R} in the first layer multiplying the weights in the second layer. Similarly, $T_3$ can be computed when one tuple in \textbf{R} appears for the first time and reused for the remaining matching tuples in \textbf{S} separately.  

However in NN literature, popular activation functions $f$ are empirically restricted to certain choices such as $sigmoid$, $tanh$ and recently $Relu$ \cite{DBLP:books/daglib/0040158}, which has been highly successful in deep learning applications primarily due to its simplicity and low overhead during optimization \cite{DBLP:books/daglib/0040158}.
It is fairly easy to confirm that both $sigmoid$ and $tanh$ are not additive functions; thus when networks utilize them, there are no opportunities to decompose and share the computation beyond the first layer.
The $Relu$ function, on the other hand, is a piece-wise linear function.
We observe that when the two terms $T_1$ and $T_2$ have the same sign, the $Relu$ function is additive. Thus, only additive functions could satisfy the requirements for exact decomposition, which limits the usage of factorized algorithms in the second layer.

Even when additive activation functions are used, it follows from Equation (\ref{eq:b0}) that computing the value before applying the activation function of a single unit requires $n_h$ multiplications and $n_h$ additions. 
After decomposition, computing Equation (\ref{eq:b}) requires summing up the results of multiplication ($w_{kj}^{(2)}f(T_1)$) and an addition (adding $T_3$). That is, it requires $n_h$ multiplications and $n_h$ additions. Furthermore, it requires another $n_h$ multiplications and $n_h$ additions to compute $T_3$ for each feature vector in \textbf{R} before it can be reused.Thus, the total number of operations required to compute a single value at the second layer is higher if we attempt to reuse the result computed from \textbf{S} and \textbf{R} respectively across layers. At higher layers, this cost is going to increase even further with a similar reasoning, which may eventually outweigh the cost savings brought by the decomposition.To summarize, reusing the computation beyond the first layer of an NN is only possible for additive activation functions, and even when this is the case, the overhead incurred by the decomposition may render any attempt to share computation at higher layers unattractive.

 \subsubsection{Computation in Backward Propagation}
\label{sec:bp}

In the backward propagation phase, the error function is $E = \frac{1}{2N}\sum_{n=1}^{N}(o^{(n)}-Y^{(n)})^2$, where $o$ is the output from the neural network and $Y$ is the target. 
Backward propagation starts from the output layer and proceeds to the lower layers. In the entire process, all the feature vectors are involved in the computation for only one time when computing the gradient of the weights between the input layer and the first hidden layer.

 $\frac{\partial E}{\partial a}$ is the gradient of error with respect to  the value before the activation function between the input and hidden layer. It is obtained via BP algorithm applying the chain rule.

{\footnotesize
\begin{align}
\underset{n_h\times d}{\frac{\partial E}{\partial w}} =&  \underset{n_h \times N}{\frac{\partial E}{\partial a}} \times \underset{N \times d}{\textbf{x}^\mathrm{T}} 
\label{eq:64} \\
=& \underset{n_h \times N}{\frac{\partial E}{\partial a}} \times {\begin{bmatrix} \underset{N \times d_S}{\textbf{x}_S^\mathrm{T}} & \underset{N \times d_R}{\textbf{x}_R^\mathrm{T}} \end{bmatrix}}
= \begin{bmatrix} \underset{n_h \times d_S}{PG_S} & \underset{n_h \times d_R}{PG_R}
\end{bmatrix} \label{eq:67}
\end{align}
}
where $PG_S =\frac{\partial E}{\partial a} \times \textbf{x}_S^\mathrm{T}$ and $PG_R =\frac{\partial E}{\partial a} \times \textbf{x}_R^\mathrm{T}$.

From Equation (\ref{eq:64}), as the result of the multiplication of the feature matrix \textbf{x}, even though we can decompose the computation into two parts as in Equation (\ref{eq:67}), there are no opportunities to explore redundancy in the computation.
The reason is that the redundancy exists among the columns of \textbf{x} due to their sharing the same part of $\textbf{x}_R$. For $PG_R$, each row in $\frac{\partial E}{\partial a}$ multiplies the corresponding column of $\textbf{x}_R^\mathrm{T}$ not $\textbf{x}_R$, i.e., {\footnotesize$[x_i^{(1)}  x_i^{(2)}  \cdots  x_i^{(N)}]^\mathrm{T}$} not {\footnotesize$[x_1^{(n)}  x_2^{(n)}  \cdots  x_d^{(n)}]^\mathrm{T}$.}

One benefit of decomposing the matrix into two parts in the backward propagation is I/O cost savings. \textbf{R} contains $n_R$ tuples, while in Equation (\ref{eq:67}), the matrix $\textbf{x}_R^\mathrm{T}$ has $N$ tuples.
We can obtain features directly from \textbf{R} and use the primary/foreign-key relationship to retrieve corresponding features from \textbf{S}. This way,  $\textbf{x}_S^\mathrm{T}$ can be populated along with the matrix  $\textbf{x}_R^\mathrm{T} $ required. In particular, as shown in Figure \ref{fig:examplenn}(b), for a feature vector $(t)$ in \textbf{R}, the matching feature vectors (2), (3) and (5) in \textbf{S} populate $\textbf{x}_S^\mathrm{T}$. Then the feature vector $(t)$ is inserted in positions corresponding to $\textbf{x}_S^\mathrm{T}$ in $\textbf{x}_R^\mathrm{T}$. Note that both $\textbf{x}_S^\mathrm{T}$  and $\textbf{x}_R^\mathrm{T}$  have $N$ rows, where $N$ is the cardinality of \textbf{S}.
Thus, instead of retrieving $N \times (d_S + d_R)$ fields in \textbf{T}, we only need $n_S \times d_S + n_R \times d_R$ fields where $n_S = N$ and $n_R < N$. Thus, {\em F-NN} can bring savings by reducing I/O cost during backward propagation even when there are no reused calculations.

To summarize, our final proposal is {\em F-NN}, which follows from algorithms {\em M-NN} and {\em S-NN} and applies the optimizations during the forward and backward propagation. {\em F-NN} reduces computational cost by removing repeated calculations through decomposition as outlined in Section \ref{sec:ff} and reduces I/O cost by avoiding reading the redundant fields in \textbf{T} as discussed in Section \ref{sec:bp}. We will experimentally evaluate such savings in Section \ref{sec:exp}.

One may notice that the proposed optimizations are compatible with popular techniques in DNN such as  batch normalization \cite{DBLP:books/daglib/0040158} (as it affects all input and applied before data enters the network) as well as Dropout \cite{DBLP:books/daglib/0040158}. Dropout can be applied either after activation at a layer or at the network input. In both cases, the linearity between the weights and the input, allows Dropout to be applied effectively with suitable bookkeeping.

\new{\subsection{Algorithm F-NN for Multi-way Joins}}

The algorithm generalizes naturally to multi-way joins. We assume the same setting involving a join of $q+1$ relations as in Section \ref{sec:multigmm}.

During forward propagation, we read features directly from tables \textbf{S}, $\textbf{R}_1$ $\ldots$ $\textbf{R}_q$. To unify notation, we denote \textbf{S} as $\textbf{R}_0$ in the sequel. Let us
focus on the computation of a single hidden unit $h_j^{(n)}$ in the first layer:

{\footnotesize
\begin{align}
\label{eq:47}
h_j^{(n)}  &=  f(\sum_{i=1}^d w_{ji}^{(1)} x_i^{(n)} + b_j^{(1)}) \\
 & = f(\sum_{i=1}^{d_{R_0}} w_{ji}^{(1)} x_i^{(n)} + \sum_{m=1}^{q}\sum_{i=d_{m-1}+1}^{d_{m}} w_{ji}^{(1)} x_i^{(n)} + b_j^{(1)}) 
\end{align}
}

We compute the weights multiplied with features from table $\textbf{R}_0$ (\textbf{S}) and then compute the weights multiplied with features from tables $\textbf{R}_1$ to $\textbf{R}_q$. After sum them along with the bias, we feed the result to the activation function to get the value of the neural network at this hidden unit. 

During backward propagation, the gradient of error with respect to the weights between the input and the first hidden layer is obtained by the BP algorithm applying the chain rule:

{\footnotesize
\begin{align}
\underset{n_h\times d}{\frac{\partial E}{\partial w}} =&  \underset{n_h \times N}{\frac{\partial E}{\partial a}} \times \underset{N \times d}{\textbf{x}^\mathrm{T}} 
= \underset{n_h \times N}{\frac{\partial E}{\partial a}} \times \begin{bmatrix} \underset{N \times d_S}{\textbf{x}_{R_0}^\mathrm{T}} & \cdots & \underset{N \times d_{R_q}}{\textbf{x}_{R_q}^\mathrm{T}} \end{bmatrix} \notag\\
=& \begin{bmatrix} \underset{n_h \times d_S}{PG_{R_0}} &\cdots& \underset{n_h \times d_{R_q}}{PG_{R_q}}
\end{bmatrix} 
\label{eq:mnn}
\end{align}}
where {\footnotesize$PG_{R_m} = \frac{\partial E}{\partial a} \times \textbf{x}_{R_m}^\mathrm{T}$, $m \in \{0, ..., q\}$}. The same optimization as in Section \ref{sec:bp} is applied to decompose the matrix multiplication into $q+1$ parts in Equation (\ref{eq:mnn}). Feature vectors can be divided into $q+1$ parts and obtained directly from $\textbf{R}_0$ to $\textbf{R}_q$ to populate $\textbf{x}_{R_m}^\mathrm{T}$ in the corresponding position. This is a way to reduce the I/O cost compared to retrieving the entire feature vectors in \textbf{T}.

\section{Experiments}
\label{sec:exp}

In this section, we present a detailed experimental evaluation of all the algorithms presented comparing their performance.
We compare the runtime performance of {\em M-GMM}, {\em  S-GMM} and {\em  F-GMM} for the case of  GMM as well as {\em M-NN}, {\em S-NN} and {\em F-NN} for the NN. There are two main parameters of interest in the underlying relations that essentially quantify the impact of normalization in terms of eliminating redundancy: the number of features in $R$ ($d_R$) and the tuple ratio of \textbf{S} and \textbf{R} ($rr = {n_S}/{n_R}$). We vary these two parameters controlling the amount of redundancy that the join introduces. In addition, we vary the number of clusters ($K$) for GMM and the number of hidden units ($n_h$) for NN . We utilize both synthetic and real datasets in our evaluation as outlined below.

\subsection{Datasets}
In order to be able to vary the parameters of interest in a controlled way to observe trends, we utilize synthetic datasets. We generate synthetic datasets for primary/foreign-key joins with a wide range of attributes in the relations involved. The parameters varied are shown in Table \ref{tab:2}, \ref{tab:3} for GMM and NN experiments respectively. We generate synthetic data sampling from multiple Gaussian distributions and add random noise in accordance with previous work \cite{DBLP:conf/sigmod/KumarNP15}.

{\small
\begin{table}[htbp]
\caption{Synthetic data dimensions for  GMM \label{tab:2}}
\centering 
\begin{tabular}{|c|c|c|c|c|c|} \hline
Experiment & $n_S$ & $n_R$ & $d_S$ & $d_R$ & $K$\\
\hline\hline
Vary $rr$ & Varied & 1000 &  5 & 5 and 15 & 5\\
\hline 
Vary $d_R$ & $10^6$ and $5 \times 10^6$ & 1000 & 5 & Varied & 5 \\
\hline
Vary $K$ & $10^6$ & 1000 & 5 & 15 & Varied \\
\hline
\end{tabular}

\end{table}
}

{\small
\begin{table}[htbp]
\caption{Synthetic data dimensions for NN \label{tab:3}}
\centering 
\begin{tabular}{|c|c|c|c|c|c|} \hline
Experiment & $n_S$ & $n_R$ & $d_S$ & $d_R$ & $n_h$\\
\hline\hline
Vary $rr$ & Varied & 1000 &  5 & 5 and 15 & 50\\
\hline 
Vary $d_R$ & $10^6$ and $5 \times 10^6$ & 1000 & 5 & Varied & 50 \\
\hline
Vary $n_h$ & $10^6$ & 1000 & 5 & 15 & Varied \\
\hline
\end{tabular}

\end{table}
}

For the real datasets, we utilize the {\em Expedia}, {\em Walmart} and {\em Movies} datasets from the Hamlet Plus Project\footnote{Available at  https://adalabucsd.github.io/hamlet.html}. We derive {\em Expedia1} dataset by joining {\em R1\_Hotels} (relation \textbf{R}) with {\em S\_Listings} (relation \textbf{S}) and  {\em Expedia2} dataset by joining {\em R2\_Searches} (relation \textbf{R}) with {\em S\_Listings} (relation \textbf{S}). For the {\em Walmart} dataset, we join {\em R1\_Indicators} (relation \textbf{R}) with {\em S\_Sales} (relation \textbf{S}) and for the {\em Movies} dataset, we join {\em R2\_movies} (relation \textbf{R}) with {\em S\_ratings} (relation \textbf{S}). The details of the datasets are available in Table \ref{tab:4}. For GMM, we use the original representation of the data (labeled {\em Not Sparse}). For NN, we use the one-hot representation of the data (labeled {\em Sparse}). 
Unless stated otherwise, we run NN training for 10 epochs and utilize a single hidden layer in our experiments. 
{\small
\begin{table}[htbp]
\caption{Data dimensions of real datasets \label{tab:4}}
\centering 
\begin{tabular}{|c||c|c|c|c|} \hline 
\textbf{Dataset} & $n_S$ & $d_S$ & $n_R$ & $d_R$ \\ [0.5ex]
\hline\hline
Expedia1(Not Sparse) & 942142 & 7 & 11938  & 8 \\
\hline
Expedia2(Not Sparse) & 942142 & 7 & 37021 & 14 \\
\hline
Walmart (Not Sparse) & 421570 & 3 & 2340 & 9 \\
\hline
Movies (Not Sparse) & 1000209 & 1 & 3706 & 21 \\
\hline
Walmart (Sparse) & 421570 & 126 & 2340 & 175 \\
\hline
Movies (Sparse) & 1000209 & 1 & 3706 & 21 \\
\hline
\end{tabular}
\end{table}}

Since the dimension in the real datasets is limited, we construct datasets derived from the {\em Expedia1} dataset with larger dimensions. These are constructed by picking tuples with a high $rr$ and increasing $d_R$ by repeating the features (as well as adding random Gaussian noise). These are depicted in Table \ref{tab:5} as {\em Expedia3} to {\em Expedia5} along with their associated characteristics.
{\small
\begin{table}[htbp]\caption{Data dimensions of augmented real datasets \label{tab:5}}
\centering 
\begin{tabular}{|c||c|c|c|c|} \hline
\textbf{Dataset} & $n_S$ & $d_S$ & $n_R$ & $d_R$ \\ [0.5ex]
\hline\hline
Expedia3 & 634133 & 7 & 2899  & 29 \\
\hline
Expedia4 & 634133 & 7 & 2899 & 78 \\
\hline
Expedia5 & 634133 & 7 & 2899 & 218 \\
\hline
\end{tabular}

\end{table}
}

To generate data for multi-way joins, we utilize the {\em Movies} dataset (relations {\em S\_ratings}, {\em R1\_users}  and {\em R2\_movies}) and inject synthetic data to relation {\em R1\_users} ($\textbf{R}_1$) keeping the size of relation {\em R2\_movies} ($\textbf{R}_2$) unchanged. In this way, we can vary the redundancy ratio between $\textbf{R}_1$ and $\textbf{R}_2$. Each time we generate a synthetic tuple for $\textbf{R}_1$, we select a random tuple from $\textbf{R}_2$, extract the key and insert a synthetic tuple on {\em S\_ratings} (\textbf{S}) enforcing any relational constraints in the process. We also vary the dimension for relation {\em R1\_users} ($d_{R_1}$) during the experiments.

\subsection{Experimental Setup}
All experiments were run on a cluster of machines with 16 Intel Xeon E5630 2.53 GHz cores, 96 GB RAM and 338 GB disk with CentOS 6.2. Our code is implemented in Python 2.7.13 using NumPy for all the matrix calculations and psycopg2 to read and write data from PostgreSQL 9.6.  The RDBMS is utilized primarily for storage of relations and all algorithm logic is implemented on top of the RDBMS. We also conducted experiments with TensorFlow in place of Numpy, as well as on GPUs instead of CPUs, and the trends are consistent with what we are presenting here; they are thus omitted due to space limitation. 

\subsection{Results on Synthetic Datasets}

\subsubsection{GMM}
\begin{figure*}[htbp]
  \centering
  \subcaptionbox{Varying $rr$}[.32\linewidth][c]{%
\includegraphics[height=4cm,width=1\linewidth]{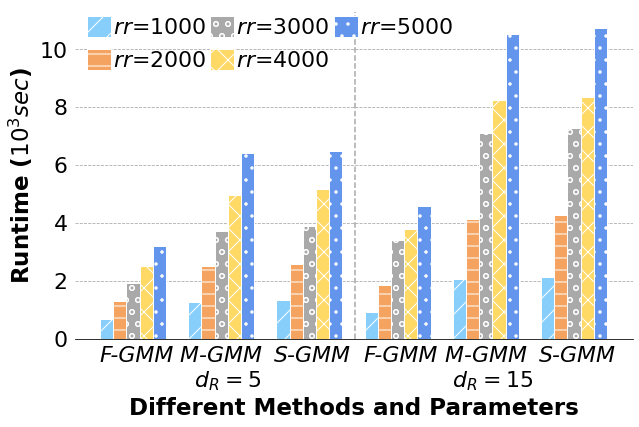}}
  \subcaptionbox{Varying $d_R$}[.32\linewidth][c]{%
\includegraphics[height=4cm,width=1\linewidth]{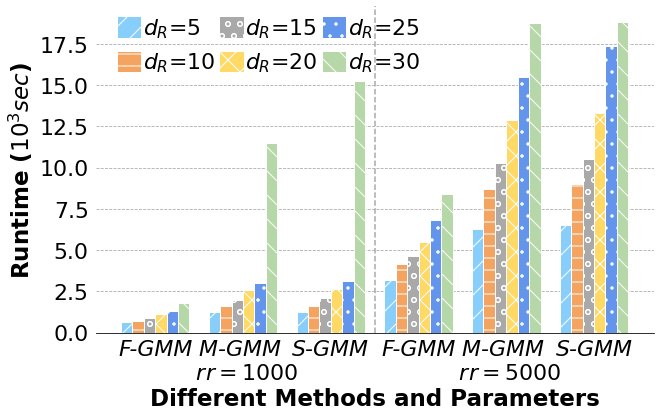}}
  \subcaptionbox{Varying $K$}[0.32\linewidth][c]{%
\includegraphics[height=4cm,width=1\linewidth]{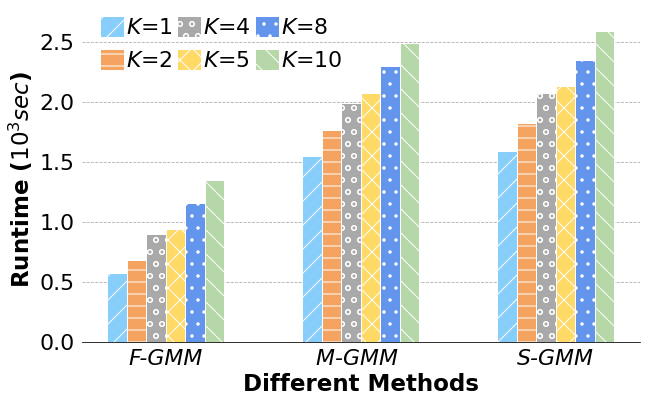}}
\caption{Performance results for GMM algorithms varying parameters of interest \label{fig:exp-gmm}}
\end{figure*}

Figure \ref{fig:exp-gmm} presents the results for the case of the GMM algorithms varying tuple ratio ($rr$) in Figure \ref{fig:exp-gmm}(a), varying the number of features in $R$ ($d_R$) in Figure \ref{fig:exp-gmm}(b) and varying the number of clusters ($K$) in Figure \ref{fig:exp-gmm}(c). In all cases, {\em F-GMM} is fastest than the other two applicable approaches. In Figure \ref{fig:exp-gmm}(a), with the increase of $rr$, the benefits of the proposed {\em  F-GMM} become increasingly larger. It can be seen that the trend will persist becoming increasingly larger as $d_R$ increases. For $d_R = 5$, {\em  F-GMM} is 2 times faster than {\em  S-GMM}, which becomes 2.4 times faster when $d_R = 15$.  In Figure \ref{fig:exp-gmm}(b), it is evident that as $d_R$ increases, {\em  F-GMM} becomes two to six and a half times faster than the other approaches for different values of $rr$. The benefit will keep on increasing as we increase $d_R$. Finally Figure \ref{fig:exp-gmm}(c) presents that {\em  F-GMM} is two to three times faster when we vary $K$ for fixed $rr$ and $d_R$. This benefit will increase as we vary $rr$ and/or $d_R$.

\begin{figure*}[htbp]
  \centering
  \subcaptionbox{Varying $rr$}[.32\linewidth][c]{%
\includegraphics[height=4cm,width=1\linewidth]{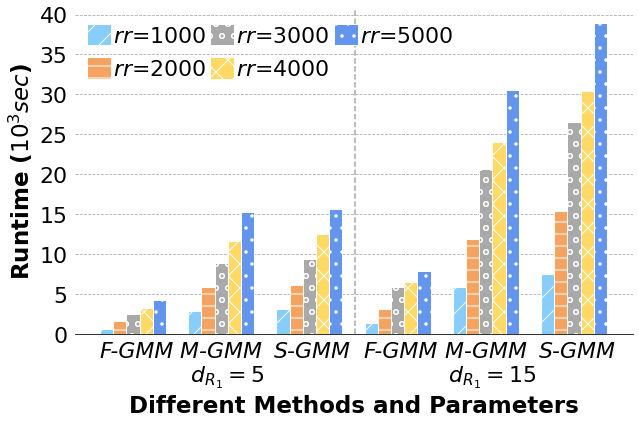}}
  \subcaptionbox{Varying $d_{R_1}$}[.32\linewidth][c]{%
\includegraphics[height=4cm,width=1\linewidth]{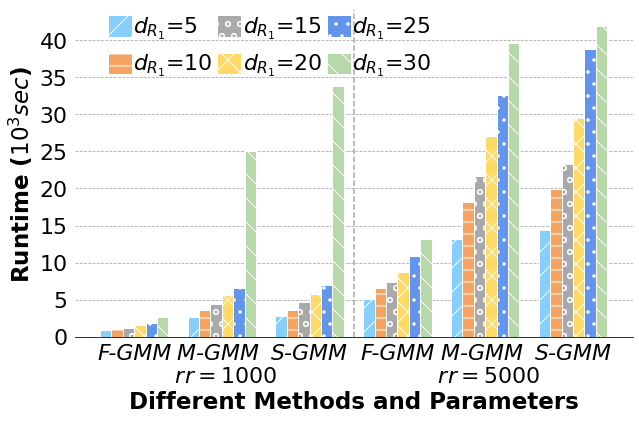}}
  \subcaptionbox{Varying $K$}[.32\linewidth][c]{%
\includegraphics[height=4cm,width=1\linewidth]{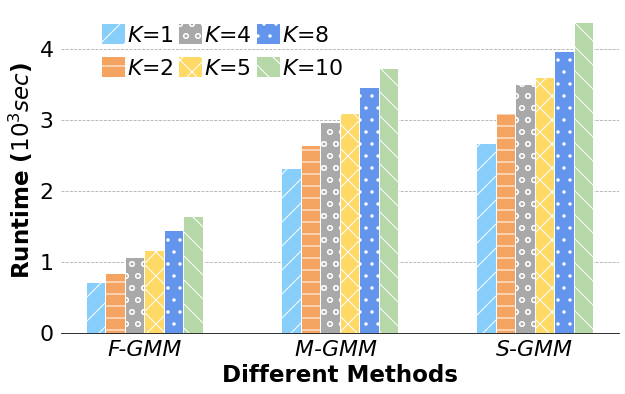}}
\caption{Performance results for GMM algorithms on multi-way joins varying parameters of interest \label{fig:exp-m-gmm}}
\end{figure*}

Figure \ref{fig:exp-m-gmm} presents the corresponding experiments for multi-way joins. The results are overall consistent, however, the benefits of our proposal for increasing $rr$ (ratio of synthetic tuples in $\textbf{R}_1$ to the number of tuples in $\textbf{R}_2$) vary from three to five times faster. As $d_{R_1}$ increases, it is three to fourteen times faster than others. Moreover, when increasing $K$, the time will be saved three to five times than alternative approaches. Compared with the results of binary joins, the optimizations introduced by {\em F-GMM} pay off as the number of joins increases.

\subsubsection{NN}

\begin{figure*}[htbp]
  \centering
  \subcaptionbox{Varying $rr$}[.32\linewidth][c]{%
    \includegraphics[height=4cm,width=1\linewidth]{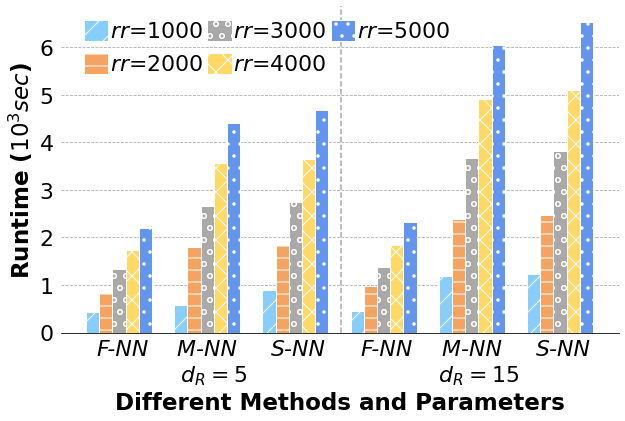}}
  \subcaptionbox{Varying $d_R$}[.32\linewidth][c]{%
    \includegraphics[height=4cm,width=1\linewidth]{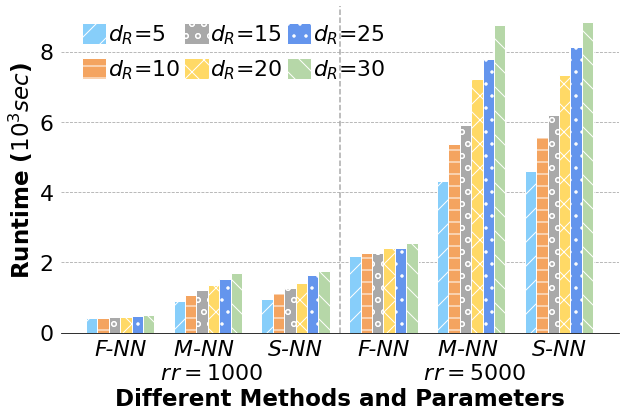}}
  \subcaptionbox{Varying $n_h$}[.32\linewidth][c]{%
    \includegraphics[height=4cm,width=1\linewidth]{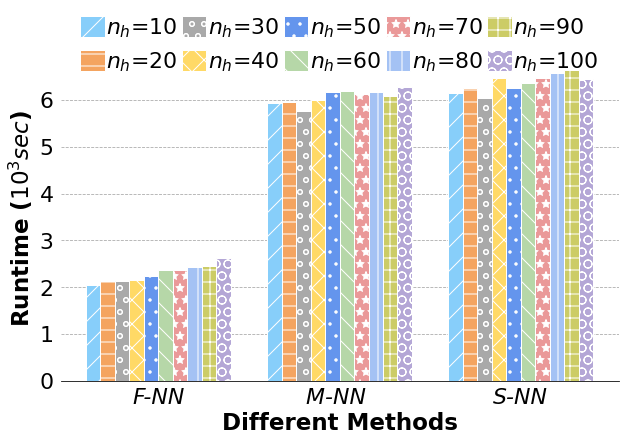}}
\caption{Performance results for NN algorithms varying parameters of interest \label{fig:exp-nn}}
\end{figure*}

{\em F-NN} gains obvious performance improvement for the case of NN. Figure \ref{fig:exp-nn}(a) presents the results when increasing $rr$. For $d_R = 5$, {\em F-NN} becomes more than two times faster. These savings will keep an upward tendency as $rr$ increases further. For $d_R = 15$, it progressively becomes three times faster, with higher benefits as $rr$ increases. Figure \ref{fig:exp-nn}(b) shows the results of a corresponding experiment varying $d_R$. With the growth of $d_R$, performance advantages vary increasingly from two to three times faster for $rr=1000$ and from 2.2 to 3.5 times for $rr=5000$. Finally, Figure \ref{fig:exp-nn}(c) reveals the results for increasing $n_h$ in the network. For fixed $rr$ and $d_R$, as $n_h$ increases, the performance benefits of {\em F-NN} vary from two to three times faster than the others. These advantages will increase as we vary $d_R$ and/or $rr$.

It is evident that for very small values of $rr$, depending on the values of $d_R$, the proposed approach may not offer performance advantages. For example, for $d_R = 5$, we observe performance benefits for values of $rr > 200$. Similarly, when $d_R = 15$, benefits start appearing for values of $rr > 50$.

\begin{figure*}[htbp]
  \centering
  \subcaptionbox{Varying $rr$}[.32\linewidth][c]{%
    \includegraphics[height=4cm,width=1\linewidth]{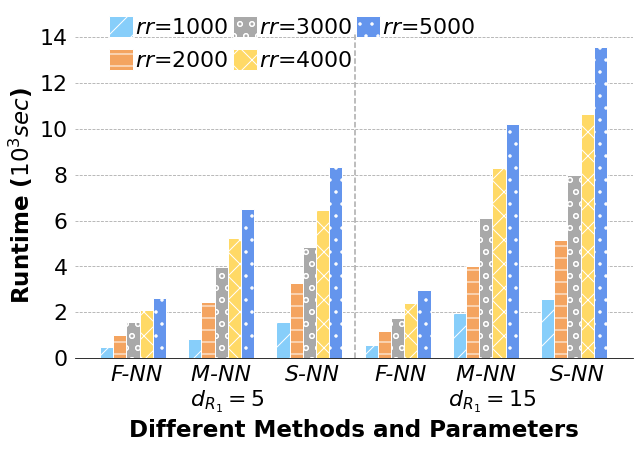}}
  \subcaptionbox{Varying $d_{R_1}$}[.32\linewidth][c]{%
    \includegraphics[height=4cm,width=1\linewidth]{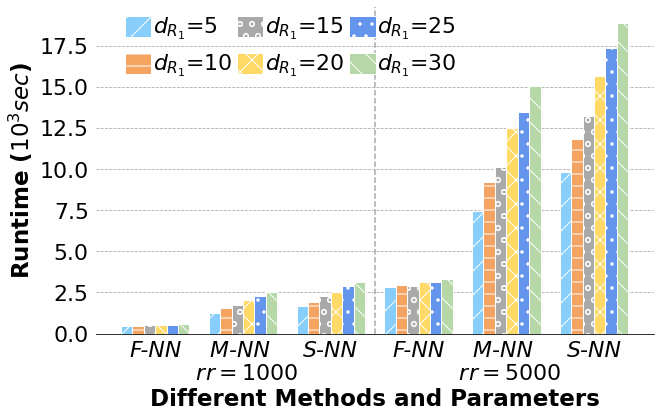}}
  \subcaptionbox{Varying $n_h$}[.32\linewidth][c]{%
    \includegraphics[height=4cm,width=1\linewidth]{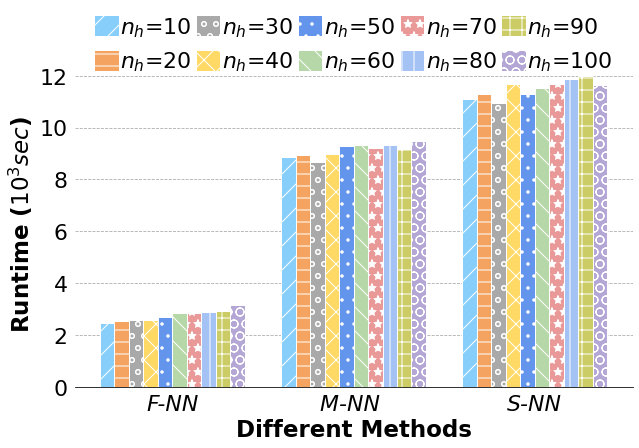}}
\caption{Performance results for NN algorithms on multi-way joins varying parameters of interest \label{fig:exp-m-nn}}
\end{figure*}

Figure \ref{fig:exp-m-nn} presents the results of the same experiments for multi-way joins. The results are consistent, however, the benefits of {\em F-NN} range from three to four times faster as we vary $rr$ and from three times faster for small values of $rr$ to six times faster for larger $rr$ as we increase $d_{R_1}$. Similarly, the benefits persist as we increase $n_h$, up to four times in Figure \ref{fig:exp-m-nn}(c). It reveals that the proposed approach offers large performance benefits especially as the number of joins increases.

\subsection{Results on Real Datasets}
The following tables present the results for the real datasets for the case of  GMM and NN respectively (all times in the tables in seconds). 
\begin{table}[htbp]
\caption{Results on real datasets for GMM } \label{tab:6}
\begin{scriptsize}
\centering 
\begin{tabular}{|c||c|c|c|c|c|} \hline
\textbf{Dataset} & {\em M-GMM} & {\em S-GMM} & {\em F-GMM} \\ [0.1ex]
\hline\hline
Expedia1(Not Sparse)  & 2140.1 & 2244.3 & 1014.2\\
\hline
Expedia2(Not Sparse) & 1221.1 & 1248.5 & 593.1 \\
\hline
Walmart (Not Sparse) & 595.9 & 602.9  & 212.1 \\
\hline
Movies (Not Sparse) & 1691.7 & 1755.8 & 514.6 \\
\hline
Expedia3 (Augmented) & 1673.5 & 1750.9 & 639.3 \\
\hline
Expedia4 (Augmented) & 6129.6 & 6311.4 & 1843.3 \\
\hline
Expedia5 (Augmented) & 23270.6 & 23375.1 & 9779.3 \\
\hline
Movies-3way & 2455.3 & 2883.1  & 715.1 \\
\hline
\end{tabular}

\end{scriptsize}
\end{table}

From Tables \ref{tab:6}, the performance benefits of {\em F-GMM} for different datasets are up to 3.4 times faster than {\em M-GMM} or {\em S-GMM}. For the same experiment in the case of {\em Expedia3} to {\em Expedia5} datasets, we can see that the benefits of {\em F-GMM} range from 2.4 to 3.4 times faster than others while as to {\em Expedia1}, they are up to 2.2 times. Extending to multi-way joins, we display the results for {\em Movies-3way}. In particular, {\em F-GMM} is 4.4 times (for {\em F-GMM}) faster compared to the materialized methods. It is evident that compared with the binary join case, the performance benefits of our approach are larger since there are more opportunities for exploiting the inherent redundancy after the join. 

\begin{table}[htbp]\caption{Results on real datasets for NN  \label{tab:8}}
\begin{scriptsize}
\centering 
\begin{tabular}{|c||c|c|c|c|c|} \hline
\textbf{Dataset} & {\em M-NN} & {\em S-NN} & {\em F-NN}  \\ [0.1ex]
\hline\hline
Walmart(Sparse) & 743.1 & 845.5 & 104.1 \\
\hline
Movies (Sparse) & 437.4 & 507.2 & 112.3 \\
\hline
Movies-3way & 890.1 & 1022.3 & 202.1 \\
\hline
\end{tabular}

\end{scriptsize}
\end{table}

Finally, Table \ref{tab:8} reveals the results for the case of NN. {\em F-NN} demonstrates 8.1 times faster execution for {\em Walmart}  dataset and 4.5 times faster execution for {\em Movies} dataset. As we have demonstrated on the synthetic datasets, the performance benefits of our approach become larger as redundancy increases. In the case of the {\em Sparse} datasets, the redundancy ratio is high after being encoded. {\em F-NN} can take advantage of this redundancy during both forward and backward propagation phases and offer superior performance benefits. These results as demonstrated using real datasets attest to the significance of our proposals given the enormous recent interest in NN and associated learning technologies in academia and industry. For the multi-way join, the benefit of {\em Movies-3way} is 3.4 times for {\em F-NN}. In general, the benefits of the proposed approach depend on the amount of redundancy the join introducing.

\section{Conclusions}
\label{sec:conc}

We propose a set of algorithms to execute popular non-linear algorithms such as Gaussian Mixture Models and Neural Networks over normalized databases. We demonstrate that by carefully reworking the basic operations of these algorithms they can be executed over normalized relational inputs offering large performance benefits over alternative approaches. 
In addition to proposing and documenting our algorithms, we present the results of a detailed experimental evaluation utilizing both synthetic and real datasets demonstrating the performance advantages on our proposals. We experimentally establish that the proposed algorithms executed over normalized relations offer very significant performance advantages that become increasingly larger as the characteristics of the underlying datasets change. In the future, it is natural to focus on other types of popular deep network architectures, including various types of autoencoders, generative models as well as deep belief networks. Finally, the mode of delivery of ML algorithms exploring factorization ideas to end users (via a UDF library or native implementations) merits further investigation.
\section{Acknowledgements}
We acknowledge the support of the Natural Sciences and Engineering Research Council of Canada (NSERC).

\bibliographystyle{abbrv}

\end{document}